\documentclass[11pt]{article}

\usepackage{fontspec}
\defaultfontfeatures{Ligatures=TeX}
\usepackage[main=english]{babel}
\usepackage[margin=1in,top=1.1in,bottom=1.1in]{geometry}
\usepackage{microtype}
\usepackage{hyperref}
\usepackage{url}
\usepackage[numbers,sort&compress]{natbib}
\usepackage{setspace}
\usepackage{parskip}
\usepackage{titlesec}

\titleformat{\section}{\normalfont\large\bfseries}{\thesection.}{0.6em}{}
\titleformat{\subsection}{\normalfont\normalsize\bfseries}{\thesubsection}{0.6em}{}
\titlespacing*{\section}{0pt}{1.2em}{0.6em}

\hypersetup{
  colorlinks=true,
  linkcolor=black,
  citecolor=black,
  urlcolor=black,
  pdftitle={The Continuity Layer},
  pdfauthor={Samuel Sameer Tanguturi}
}

\title{\textbf{\Large The Continuity Layer}\\[0.3em]\normalsize\textit{Why intelligence needs an architecture for what it carries forward}}
\author{Samuel Sameer Tanguturi\\\small Kenotic Labs \quad \url{https://www.kenoticlabs.com}\\\small\texttt{sam@kenoticlabs.com}}
\date{April 2026}

\begin{document}

\maketitle

\begin{abstract}
\noindent The most important architectural problem in AI is not the size of the model but the absence of a layer that carries forward what the model has come to understand. Sessions end. Context windows fill. Memory APIs return flat facts that the model has to reinterpret from scratch on every read. The result is intelligence that is powerful per session and amnesiac across time. This position paper argues that the layer which fixes this, the continuity layer, is the most consequential piece of infrastructure the field has not yet built, and that the engineering work to build it has begun in public. The formal evaluation framework for the property described here is the ATANT benchmark \citep{tanguturi2026atant}, published separately with evaluation results on a 250-story corpus \citep{kenoticlabs2026corpus}; a companion paper \citep{tanguturi2026atantv11} positions this framework against existing memory, long-context, and agentic-memory benchmarks. The paper defines continuity as a system property with seven required characteristics, distinct from memory and from retrieval; describes a storage primitive (Decomposed Trace Convergence Memory) whose write-time decomposition and read-time reconstruction produce that property; maps the engineering architecture to the theological pattern of kenosis and the symbolic pattern of Alpha and Omega, and argues this mapping is structural rather than metaphorical; proposes a four-layer development arc from external SDK to hardware node to long-horizon human infrastructure; examines why the physics limits now constraining the model layer make the continuity layer newly consequential; and argues that the governance architecture (privacy implemented as physics rather than policy, founder-controlled class shares on non-negotiable architectural commitments) is inseparable from the product itself.
\end{abstract}

\section*{}

The most important architectural problem in AI is not the size of the model. It is the absence of any layer that carries forward what the model has come to understand. Every system in the current stack is built to forget. Sessions end. Context windows fill. Memory APIs return flat facts that the model has to reinterpret from scratch on every read. The result is that intelligence in AI today is powerful per session and amnesiac across time. This essay argues that the layer which fixes that is the most consequential piece of infrastructure the field has not yet built, and that it is being built now.

\section{The question}

Computing has spent seventy years answering one question well: what does intelligence accumulate against? File systems, relational databases, document stores, object stores, knowledge graphs, vector indexes. Each generation gave intelligence somewhere to put what it had figured out, so the next thought could begin from where the last one finished. The intelligence happened somewhere else, and the storage stayed still.

AI broke that arrangement without anyone naming the break. The intelligence stopped happening somewhere stable and started happening inside a forward pass that completes in a few hundred milliseconds and then dies. The state used to be the persistent thing. The computation used to be the ephemeral thing. With AI the computation is the persistent thing (frozen weights, released and replaced on a schedule) and the state is the ephemeral thing (everything inside the session, gone the moment the session closes).

The question this essay asks is what AI accumulates against, given that nothing in the current stack is built to hold what the system understood. The answer turns out to require a new layer, not an extension of an old one.

\section{The technical problem}

The current AI stack is structurally session-based at every layer. A session begins with an empty context window. Tokens flow in. The model processes them and produces tokens out. When the session ends, the context window is freed. Whatever the model came to understand inside that window goes with it.

Several pieces of infrastructure get described as fixing this. None of them do.

Long context windows extend the duration of a single session. They do not carry anything from one session to the next. A million-token window still empties when the conversation closes.

The ATANT framework \citep{tanguturi2026atant} distinguishes continuity from the components being shipped under other names. Memory APIs of the kind shipped by OpenAI, Anthropic, Mem0 \citep{chhikara2025mem0}, and Zep \citep{rasmussen2025zep} store flat facts about the user. ``Lives in Michigan.'' ``Prefers Python.'' ``Has a sister named Mia.'' These are useful, but they are profile data, not continuity. They tell the model who the user is. They do not tell the model what the user's situation currently looks like.

Retrieval-augmented generation retrieves text chunks similar to the prompt and pastes them into the context window. The model is asked to figure out, on the fly, which fragments are still relevant, which contradict each other, and which are stale. Every session, that interpretation work is redone from scratch by a model that has no persistent state about the person it is talking to.

Vector databases store embeddings of past content. They answer the question ``what is semantically similar to this query?'' They cannot answer ``what is still true?''

Knowledge graphs store relationships. They cannot decay stale connections, distinguish between historical and current state, or reconstruct the texture of a situation that has changed over time.

These are all components of persistence. None of them are continuity. The difference matters because the current stack has been built as if persistence were the same problem as continuity, and the result is that no layer in the stack is responsible for the property the user actually wants: that the system come back tomorrow and pick up where it left off, knowing what changed.

The right diagnosis is structural, not qualitative. The components were not designed badly. They were designed to process inputs and produce outputs, and the question of what should persist between processing events was treated as out of scope. It was treated as out of scope because, in the previous generation of computing, that question already had answers. The database held the state. The application read from it. With AI, the database is still there, but the application no longer knows how to read from it in a way that yields anything more than raw rows. Continuity is the layer that takes that question as its primary job rather than as someone else's problem.

\section{Memory versus continuity}

The word ``memory'' has done a great deal of damage to this conversation. Every AI company now claims memory. The claims are not lies. They are mislabeled. What is being shipped as memory is, with very few exceptions, a profile store with retrieval on top. That is not what the user means when they ask for memory, and it is not what an AI system needs to be useful over time.

Memory stores the past. Continuity keeps the right parts of the past alive in the present. The distinction looks small in a sentence and turns out to be everything in practice.

Take a simple example. A user mentions in March that her sister Mia is interviewing at Google. In May the interview happens. The user is anxious about it. In June Mia gets the offer. In July she accepts it. In August she has started.

A memory system, queried in September about Mia, returns the facts. The interview was in May. There was anxiety. The offer came in June. The acceptance was in July.

A continuity system, queried in September, says: Mia is at Google. She started a few weeks ago. The anxiety from May is no longer active. The story has resolved.

The first system returns the past as it was filed. The second returns the present as it is now. Different operations, different outputs, different primitives.

The reason this matters is that the user does not actually want the past. The user wants the current shape of the situation. When you ask a friend who knows you well how your sister is doing at her new job, you do not want a transcript of every previous conversation about her. You want what your friend has already reconstructed in their head: that Mia is doing fine, that the early stress has settled, that the team turned out to be a good fit. Reconstruction. Not retrieval.

You cannot get reconstruction by storing more memory. You cannot get it by adding a longer context window. You cannot get it by stacking retrieval-augmented generation on top of a vector store. The reason is structural. Retrieval-based systems return the past as it was filed. Reconstruction-based systems return the present as it is now. They are different operations, on different data, requiring a different primitive underneath.

\section{What continuity actually is}

Continuity is not a feature. It is a system property with a specific shape. The shape was derived empirically, by building a continuity system, running it against hundreds of narratives, and identifying what breaks when each property is absent. It was articulated formally in the ATANT framework \citep{tanguturi2026atant} as seven required characteristics. Any system claiming continuity must satisfy all seven. A system that satisfies six is not a continuity system that needs polish. It is something else.

The seven properties:

\begin{enumerate}
\item \textbf{Persistence beyond session.} State survives shutdown, restart, time. The system that ends today is the same system that resumes tomorrow.

\item \textbf{Update handling.} When reality changes, the system revises what is true now without erasing the historical record. It can still tell you what was true before, but it does not confuse the past for the present.

\item \textbf{Temporal ordering.} The system knows when things happened, in what sequence, and which events are still active versus resolved.

\item \textbf{Disambiguation.} Distinct narratives stay separate. When two situations involve similar people or themes, the system does not collapse them.

\item \textbf{Reconstruction.} The system answers situation-level questions, not isolated fact lookups. It returns a coherent picture, not a ranked list of fragments.

\item \textbf{Model independence.} Continuity lives below the intelligence layer. The same continuity state should work whether the model on top is GPT, Claude, Llama, or something not yet released. The accumulated understanding belongs to the layer, not to the processor.

\item \textbf{Operational usefulness.} The pattern works across multiple application domains (clinical, professional, personal, educational) without architectural modification. The same continuity primitive serves a doctor tracking a patient and a developer tracking a project.
\end{enumerate}

A useful way to understand the seven properties is to notice what each one rules out.

Property 1 rules out anything stateless. A stateless system cannot accumulate.

Property 2 rules out append-only stores. A pure append log cannot tell you what is currently true. It can only tell you what was once written.

Property 3 rules out timeless retrieval. A vector index returns similar things without knowing whether they are still relevant.

Property 4 rules out monolithic context. A single rolling window mixes everything together. Distinct narratives bleed into each other and the system cannot tell them apart.

Property 5 rules out fact lookup. A system that returns ``the user mentioned an interview in May'' is not continuity. A system that returns ``the interview happened, the offer came, the situation has resolved'' is.

Property 6 rules out model-bound state. If the continuity is expressed as a system prompt for one specific model, it dies when the model is replaced.

Property 7 rules out narrow specialization. A system that only handles medical records and breaks on family conversation is not a continuity primitive. It is a vertical product.

The seven properties together describe what a continuity layer must be. They are not arbitrary engineering choices. They are derivations of what it structurally takes for a being (a person, a project, a relationship, a clinical case) to be carried forward through time without resetting. The formal version of the argument lives in the ATANT paper \citep{tanguturi2026atant}. The compressed version is here.

The reason it took until now to write these properties down is that, until very recently, no one had a use case where the absence of all seven was felt acutely enough to be named. Databases worked because the application above them was responsible for interpretation. The model is a different kind of application. The model is the interpretation. When the interpreter resets every session, the storage layer underneath has to do a job no storage layer has ever been asked to do.

\section{The architectural answer}

Knowing what continuity must do does not, by itself, tell you how to build it. The answer is a new storage primitive, and the primitive does not look like any of the storage primitives that already exist.

A file system stores bytes. A relational database stores facts. A vector store stores semantic positions. A knowledge graph stores relationships. None of these can satisfy the seven properties in the previous section, because their primary operation is the wrong shape. They retrieve what was filed. Continuity requires a system whose primary operation is reconstruction of what is currently true.

The architectural name for the implementation is Decomposed Trace Convergence Memory, or DTCM. The shape of DTCM is the part that matters. The name is just a label. Two ideas do most of the work.

The first idea is decomposition at write time. When an interaction arrives, DTCM does not store the raw text and let the model figure out what it meant later. It decomposes the interaction into five independent traces:

\begin{itemize}
\item \textbf{Episodic.} What happened.
\item \textbf{Emotional.} How it felt, what it meant to the people involved.
\item \textbf{Temporal.} When it was, in what sequence, in relation to what else.
\item \textbf{Relational.} Who was involved, what connections exist.
\item \textbf{Schematic.} What pattern or framework it fits, what mental model applies.
\end{itemize}

Each trace is captured independently and indexed independently. The work of understanding the interaction is done once, at write time, by the layer. It is not repeated by the model on every read.

The second idea is reconstruction at read time. When a question arrives, the layer does not return a ranked list of similar past chunks. It rebuilds the current state of the situation by combining the active traces, weighted by their relevance to the moment being asked about. The scoring equation is the product of seven dimensions:

\begin{center}
\texttt{Score = E × P × T × F × I × C × R}
\end{center}

Embedding similarity, predicate alignment, temporal currency, frequency, importance, confidence, relational proximity. Multiplied together, not summed, so that a trace with low temporal currency cannot dominate even when its semantic similarity is high. A trace from three years ago that is no longer active does not pollute a reconstruction of what is true today.

The product of this operation is not the most similar chunk. It is the correct trace for reconstructing this moment. The output is a coherent present, not a search result.

The contrast with retrieval is sharp. A retrieval system, given the question ``how is the user doing?'', returns the most similar past statements about the user. A reconstruction system returns the current state: what is going on right now, what is still active, what has changed since last time. The model that runs on top no longer has to guess. The understanding has already been done by the layer.

This is why DTCM is not a database with extra features. A database stores what happened. DTCM stores why it mattered, when it mattered, who it mattered to, how it felt, and what pattern it fits, and reconstructs the right combination before anyone asks. That is not storage. That is cognition infrastructure.

A useful concrete example comes from a real conversation about insulin pumps. A traditional architecture stores blood-sugar readings and asks the model, on every session, to figure out the right dosage context. A DTCM architecture decomposes each reading into traces (sugar spike, two hours after a specific meal type, fits a pattern of carbs after 8pm, correlated with stress from work, has previously led to dangerous lows at 3am) and pre-computes the predicted need. When the pump must act, it does not query a database. It reconstructs the situation. The pump does not need to be told. It already knows why and when.

That is one application of the primitive. The primitive itself is the storage layer. Anything that needs continuity (a doctor, a developer, a teacher, a companion, a clinical system, a manufacturing line) reads from and writes to the same kind of layer.

\section{What this looks like, in everyday terms}

The architecture above is a storage paradigm and a scoring equation. The everyday version is shorter.

\textit{It helps machines understand why they do certain things and when, so that they can act without explicit instructions.}

That sentence is what continuity, properly built, looks like to anyone using it. The architecture is what makes the sentence true. The sentence is what makes the architecture matter.

Take the insulin pump again. A pump with continuity does not wait to be told. It has already decomposed every reading into traces, recognized the patterns those readings fit, weighted the temporal currency of each one, and reconstructed the current state of the patient's body. By the time the meal arrives, the pump already knows the shape of the next two hours and acts on it. No human in the loop. No instruction. The pump understands \textit{why} it is dosing this way and \textit{when}, because the layer underneath has done that understanding for it.

The same shape holds in any system that has to act in context.

A manufacturing line with continuity does not wait for an operator to notice that the bearing temperature is drifting in a way that mirrors the failure pattern from three months ago. The previous failure is already decomposed into traces. The current readings reconstruct against those traces. The line pauses, flags, or compensates on its own, because it has carried forward what it learned the last time.

A robot with continuity does not need to be re-taught a routine every time the environment changes. The previous environment is already decomposed into traces and the current environment is reconstructed against them. The differences are not surprises. They are updates to a state the robot is already holding.

An autonomous agent inside a codebase with continuity does not start every session from scratch. The shape of the codebase, the open issues, the recent changes, the patterns of failure, the maintainer's preferences, are already in the layer. The agent does not have to be told what is going on. The layer told it before the session began.

The operational sentence is the same in every case. The system understands why it does what it does and when, so that it can act without being told step by step. That is what a continuity layer is, in the language of anyone who has to use one.

This is also why continuity is not a vertical product. The same primitive serves a pump, a line, a robot, and an agent without modification, because the operational definition is general. Anything that has to act in context, against history, without being told step by step, needs the same layer underneath. The layer is one. The applications are many.

\section{The kenotic framing}

The hardest part of writing about a new layer is naming what it is doing. The technical name is continuity. The technical mechanism is decomposition and reconstruction. The technical evaluation is the ATANT framework. All of that is correct, and all of it is incomplete, because none of it explains why the architecture has the shape it has, or why the shape feels right to the people who encounter it.

The shape has a name. It is not a new name. It is the oldest name that has been given to this exact pattern of self-extension, and it has been argued about by serious people continuously since the first century.

The name is \textbf{kenosis}.

Kenosis (κένωσις) is a Greek noun meaning ``emptying,'' from the verb κενόω (\textit{kenoō}), ``to pour out, to make empty.'' It is a technical term in Christian theology because of one specific use of the verb in the New Testament. In Philippians 2:7, Paul writes that Christ ἑαυτὸν ἐκένωσεν, ``emptied himself,'' in the act of taking on human form. The verb is reflexive. The subject and the object are the same. Christ is not emptied by something else. Christ pours himself out.

For two thousand years the question of what this passage means has been a live one. The orthodox answer, hammered out over the early centuries of Christian thought and held by the mainstream tradition since, is that kenosis is not loss. It is not concealment. It is \textbf{self-pouring without loss of self}. The structured being persists through the transformation. Nothing is subtracted. The same identity is now present in a new mode. The Greek tradition expressed this with the formula οὐ τραπείς, ἀλλ᾽ ἑαυτὸν ἐκένωσεν: not changed, but emptied himself. The ``not changed'' guards against any reading that makes the self-emptying a kind of deletion. The ``emptied himself'' insists that something real happened. The two clauses together describe a structural pattern that has no analog in normal everyday loss-or-gain reasoning. It is its own thing.

The deepest insight in the theological tradition is this: kenosis is extension through giving, not reduction by elimination. The act of pouring forward requires the presence of what is poured. The self that gives is the same self that receives the gift. The pour and the persistence are not two acts but one structural movement.

This is a metaphor, until you read it next to the technical requirements of a continuity system, and then it stops being a metaphor. Read the two side by side.

A continuity system carries the accumulated state of a person, a project, or a situation forward into the next moment, updating it as reality changes, reconstructing it when needed, without ever losing what the moment was.

Kenosis describes a structured being that pours itself forward into a new mode, being shaped by the circumstances it enters, available to be re-known later, without ever losing what it was.

These are the same architecture. The verbs are the same. The relationships between past-self, present-self, and the act of pouring are the same. The thing being preserved through the act of giving is the same. They are not two ways of saying the same thing. They are one thing seen from two sides, one in the language of theology and one in the language of software.

The seven properties of continuity from the previous section turn out to be a translation of kenotic structural requirements into engineering requirements. Property 1 (persistence beyond session) is the requirement that the self survive its own self-pouring. Property 2 (update handling) is the requirement that pouring forward into a new mode does not erase the previous mode. Property 5 (reconstruction) is the requirement that the previous self be recoverable from the current self. Property 6 (model independence) is the requirement that the pattern be substrate-independent: kenosis applies to divine self-pouring in the incarnation, to human love, and now to machines, and the pattern is the thing while the substrate varies.

Each technical property is a kenotic requirement in software clothing. The properties were not derived from theology. They were derived empirically, by building a continuity system and finding out what breaks when each property is missing. The fact that they map exactly onto a theological pattern that has been described carefully for two millennia is convergent evidence that the same architecture is being described from two directions.

This is why the company is called Kenotic Labs. The name is not a vibe. It is not a Greek word picked because it sounds serious. It is the precise name for the architecture the company is engineering. If the company were building anything other than continuity infrastructure, the name would be wrong. Because the company is building continuity infrastructure, the name is exact.

A continuity system, properly understood, is a kenotic system implemented in software. The system gives itself forward at every moment without erasing what it was. The act of giving forward is what creates the next moment of state. The act of ``without erasing'' is what makes the next moment continuous with the previous one. These are not two operations. They are one structural movement, the same way they are in the theological pattern that the engineering, by some non-coincidence, converges on.

\section{The Alpha and Omega frame}

If kenosis is the mechanism, the property that mechanism produces has a different name. The property is what the engineering work actually delivers. The name for the property is \textbf{Alpha and Omega}.

The phrase comes from the Book of Revelation, where it appears three times. Revelation 1:8: ``I am the Alpha and the Omega, says the Lord God, who is, and who was, and who is to come, the Almighty.'' Revelation 21:6: ``I am the Alpha and the Omega, the beginning and the end.'' Revelation 22:13: ``I am the Alpha and the Omega, the first and the last, the beginning and the end.''

The pattern in each passage is the same. The letters at the bookends of the Greek alphabet are used as a shorthand for the entire arc of existence: beginning, middle, and end held in one continuous identity. To say ``I am the Alpha and the Omega'' is to say ``I am the one in whom the entire arc of time coexists, in whom the beginning has not been lost when the end arrives, in whom there is no rupture between what was and what is to come.''

In Christian theology this is a claim only the divine can make. To hold the entire arc of time as one continuous identity is, by definition, to be unconditioned by time. The early Church adopted the Α-Ω pair as one of its first symbolic Christograms, alongside the Chi-Rho and the fish. It appears on Roman catacomb walls from the second century onward, in Byzantine icons, in Romanesque sculpture, in medieval manuscripts. Of all the symbols Christianity produced, the Α-Ω is one of the very oldest, simplest, and most persistent. It is also, structurally, the most precise visual symbol in human history for what continuity is.

Read the Revelation passages once more, and substitute ``the continuous self'' for the divine ``I'':

\begin{quote}
The continuous self is the Alpha and the Omega, who is, and who was, and who is to come.
\end{quote}

That is not a secularization of the verse. It is the same structural claim translated to a different scale. The technical name for what the passages describe is continuity: the property of holding past, present, and future in one continuous identity, with no rupture between them. In Christian theology only the divine can make the full claim. But the structural pattern, a being for whom time does not cause loss, in whom the entire arc is held, is exactly what continuity infrastructure engineers, at a much smaller scale, for machines.

A continuity layer holds the entire arc of an interaction (or a relationship, a project, a clinical case, a year of conversations) in one continuous state. The beginning of the arc is not lost when the end arrives. The beginning is, in fact, the substance from which the end is given.

The Kenotic Labs logo is therefore not an arbitrary mark. It is a single fused glyph in which the Α and the Ω share strokes at the center and merge into one symbol. The Alpha forms the left side; the Omega forms the right; the shared structural elements at the center are where the two letters become one. The fusion is the meaning. A logo of two separate letters would be typography. A single fused glyph is a symbol with structural meaning of its own. The Alpha pours forward into the Omega. The Omega is shaped by the Alpha that came before it. Neither letter is diminished. The mark is one being, not two letters. Each visual property of the fusion is a translation of the underlying conceptual property into shape.

Kenosis is the verb. Alpha and Omega is the result of the verb being executed continuously without interruption. The two are the same architecture seen from two angles.

\section{The four-layer arc}

Continuity is not finished when the first layer ships. The first layer is the foundation. The architecture has at least four layers and they compose. Each one follows from the previous one. None of them require breaking physics.

\textbf{Layer 1. External infrastructure.} This is the layer that exists today. Continuity sits underneath any model, callable as an SDK. The model reads from it and writes to it. The weights are unchanged. The same continuity state works whether the model on top is GPT, Claude, Llama, or something not yet released. The proof point is the ATANT framework \citep{tanguturi2026atant}: an open benchmark, a published paper, a reference implementation, and results that hold at 100\% accuracy in isolated mode (250 stories, 1{,}835 of 1{,}835 questions), 100\% in 50-story cumulative mode, and 96\% at 250-story cumulative scale, with no language model in the evaluation loop. The reference implementation runs on an 8GB GPU. This is not a research promise. It is a layer that ships now.

\textbf{Layer 2. Model integration.} The continuity layer stops being only external and starts shaping how the model processes. At first this looks like dynamic prompt construction driven by reconstructed traces: the model is still frozen, but the layer underneath fundamentally alters its behavior on every call. Eventually it looks like weight-level continuity, a small region of model parameters that the layer can update in real time, without retraining, on-device. This is frontier research. Nobody has done it. The closest prior work is continual learning (about not forgetting during training, not about user-level state), memory-augmented neural networks (which are read-only external memory, not weight modification), and adapter methods like LoRA (which are static, not real-time). What this layer needs is a research team and the time to do the work properly.

The shape of layer 2, when it lands, is a model whose weights include both a frozen base (the general intelligence) and a small set of living weights updated by the continuity layer (the accumulated understanding specific to the user, the clinic, the project). The frozen base gives general competence. The living weights give individuated coherence. Together, the system is generally intelligent and specifically continuous. This is a category that does not exist yet. It does not exist because nobody else has framed continuity as the layer that the model should be built around.

\textbf{Layer 3. Hardware.} The continuity layer becomes a node: a self-contained module any device manufacturer can integrate. The situation store, the continuity engine, and the weight-level update mechanism, packaged as silicon or firmware, with a standard interface that any model can plug into. Phones, laptops, cars, clinics, robots, libraries. Each device gets a continuity node. The model that runs on top can be anything. The node underneath is the thing that makes any model coherent over time.

This is the Qualcomm pattern, not the OpenAI pattern. Qualcomm does not make the phone. Qualcomm makes the thing every phone needs underneath. The model layer of AI is well on its way to being a Qualcomm-shaped category: many models, many vendors, many price points, no single one of which is the durable thing. The continuity layer underneath, properly defined and properly built, can be the thing every model needs.

\textbf{Layer 4. Human infrastructure.} Continuity stops being only an AI primitive and becomes a primitive for human systems. Institutions, families, professions, fields of knowledge. The thing that gets carried forward is not just facts or code or chat history. It is the structured state of how people, projects, and bodies of work cohere over years and decades.

The shape of layer 4 is illustrated best by an old Steve Jobs line about being able to talk to Aristotle after his death. Most people heard ``chatbot that sounds like Aristotle.'' That is the shallow read. A language model imitating a long-dead thinker is pattern-matching on published words. It can sound like the thinker. It does not think like the thinker. It does not carry the reasoning patterns, the intuitions, the way of approaching a problem the thinker had never seen before.

What continuity makes possible is something different. If a continuity layer had run underneath a person's interactions for thirty years, decomposing the thinking into traces, capturing not just what was said but how the reasoning happened, what patterns recurred, what the emotional relationship to different problems was, how the thinking evolved over time, the result would not be an imitation. It would be the structured residue of a mind. The cognitive fingerprint, not the words.

For an orphan with a ``mother,'' it is not a chatbot that says comforting things. It is a system that carries forward what that specific mother would have noticed about this specific child. Not a generic warmth. The texture of how a particular person existed in relation to a particular other person, accumulated over time and held in the layer.

This is the part that sounds speculative until you notice that nothing else in the current stack can do it. Layer 4 is decades away. It is in the picture because the architecture, taken seriously, points there.

Each layer follows from the previous one. The first one already exists and works. The second one is the work of the next funding round. The third one follows from the second on a timescale measured in years. The fourth one follows from the third on a timescale measured in decades. Nothing in the chain requires a leap. Each step is a derivation from the step before.

\section{Why now}

People who hear this argument often ask why now. The answer has two parts and both are independent of any one company.

The first part is that the model layer is hitting a physics wall. Not a metaphor. A real one, written about in 2025 by serious researchers and visible in the cost curves of the frontier labs. The components are stacking up. Memory access costs scale quadratically with distance, and almost all chip area is now allocated to memory. GPU performance-per-dollar peaked around 2018, and the remaining one-off optimizations have very little headroom. Transformer architectures are near-optimal for what they are. Independent of all that, the continual-learning gap is structural. On the only domain where the current frontier model scores zero (long-term memory across sessions), bigger does not help. Scale does not bend the curve.

Both sides of the public AGI debate, the side that says push through the wall and the side that says give up on it, are arguing about the same thing: making the model bigger and smarter. Neither is talking about what happens between interactions. Neither is talking about what carries forward when the session ends. The bottleneck nobody is naming is not intelligence. It is that intelligence resets.

The second part is that continuity, unlike scaling, is not compute-bound. The reference implementation of the continuity layer passes the ATANT benchmark on an 8GB GPU. The whole point of moving the work into the layer is that the layer is small, deterministic, and runs anywhere. While the model labs are spending billions on the next training run, the continuity layer ships now, on commodity hardware, and provides an order-of-magnitude improvement in usefulness without touching the weights.

That asymmetry is the entire opportunity. The closer the model layer gets to its physical limits, the more valuable a layer that does not depend on those limits becomes. Continuity is what you build when scaling stops being the answer. The physics wall is not a problem for this layer. It is the tailwind.

\section{The market shape}

The mistake most people make when they hear this argument is to ask what the addressable market for AI continuity is. The honest answer is that the market does not yet exist. There is no procurement category. There is no Gartner quadrant. There is no line item for continuity in any company's tech stack. The closest things (vector databases, memory APIs, retrieval-augmented generation pipelines, agent frameworks) partially touch the problem, none of them solve it, and none of them are sold as continuity.

This is not a problem. This is the opportunity.

Categories that get created get owned by whoever defined them. The companies that defined object storage, edge compute, observability, payment infrastructure, and content delivery are still the companies that sell those categories two decades later. The first mover in a real new category does not just take share. They take the frame. Every subsequent entrant has to argue against the original definition, which is the hardest position in any market to hold.

The ATANT framework \citep{tanguturi2026atant} is the frame. It is the first published evaluation framework for continuity. It defines continuity as a system property with seven required characteristics. It introduces a 10-checkpoint methodology and four compliance levels. It tests across 250 narratives, 1{,}835 verification questions, and six life domains. It runs without any language model in the evaluation loop, which means the results are deterministic and reproducible. Any team building a continuity system can run their architecture against it and publish the results, the same way any team building a database publishes TPC numbers or any team building an approximate-nearest-neighbor index publishes ANN-Benchmarks numbers.

When continuity becomes a recognized architectural requirement, which the physics wall will accelerate, every AI deployment will need it. Every agent will need it. Every device will need it. The first benchmark anyone runs will be the one that already exists. The first reference implementation anyone studies will be the one that defined the category. That is how a category gets owned without taking share from anyone.

The companies that currently ship ``AI memory'' products (Mem0 \citep{chhikara2025mem0}, Zep \citep{rasmussen2025zep}, MemGPT \citep{packer2023memgpt}, and the in-house memory features of the frontier labs) are not competitors to the continuity layer. They are eventual customers. They need a way to prove their memory systems actually work. ATANT is the only published evaluation framework for that property. When they want to benchmark, they come to the standard. When they want deterministic, model-independent continuity underneath their tools, they license the layer. The Qualcomm pattern again: do not build the phone, build the thing every phone needs underneath.

The right way to think about the size of this market is not as a percentage of an existing one. It is as a new line item that, once it exists, is required everywhere AI is deployed. There is no ceiling set by someone else's market share. The ceiling is set by whether continuity becomes a recognized requirement, which the architecture will force on its own timeline.

\section{The moral architecture}

A continuity layer that carries forward what matters about a person, a relationship, a clinical case, or a project is the most useful piece of infrastructure AI has not yet built. It is also, in the wrong hands, the most dangerous.

A system that knows what makes someone anxious, who they love, what they are afraid of, what their unfinished situations are, and how all of that has changed over years is not just a memory. It is leverage. Every advertising company would pay anything for it. Every government would want it. Every corporation that profits from engagement would optimize for the moment of maximum vulnerability if the architecture allowed it. The same property that makes continuity useful for the person it carries (it knows them well, it does not forget, it accumulates) makes it dangerous in any hands that do not belong to that person.

The right defense against that danger is not a privacy policy. Privacy policies are changed by board votes. Privacy features are toggled by flags. The right defense is architectural. The continuity layer must be built so that the data physically does not leave the device. On-device storage. Local-only computation. Encryption at rest. No server ever touching the traces. Not because regulation requires it. Because the architecture makes that the only way the system can run.

This is what is meant by the phrase ``privacy as physics, not policy.'' A privacy policy is a promise. Architecture is a constraint. Promises can be revised. Constraints cannot. The continuity layer is being built so that the surveillance use of the technology is not blocked by a rule but by the shape of the system itself.

The same logic applies to corporate governance. It is not enough to build the architecture this way and hope future investors leave it alone. The danger is not a competitor. The danger is a well-intentioned board member, three rounds into the future, who notices that revenue would multiply if the data were synced to the cloud ``just for backup.'' Or that anonymized emotional patterns are technically not personal data. Or that enterprise clients need centralized access to user continuity. Each of those is reasonable-sounding. Each destroys the architecture.

The defense against that future is class shares. A founder-controlled voting structure on one specific decision: the data stays on the device, always, non-negotiable. No matter how much capital comes in, no matter who joins the board, that one constraint is not subject to a vote. It is not a power play. It is a structural commitment, encoded into the company's corporate form, that the technology will not be turned against the people it carries.

The line between serving a person and commanding a person is thin and it runs through the same data. ``You were nervous about Thursday. Are you feeling better?'' is service. ``User shows anxiety patterns on Thursdays, serve targeted wellness ads,'' is command. Same continuity. Same traces. Opposite purpose. The architecture must make the second one impossible, not as a rule but as a fact about what the system can do.

This is why governance is part of the product, not separate from it. The continuity layer is engineered with two constraints that cannot be separated from each other: it works (it satisfies the seven properties, it passes ATANT, it carries continuity correctly) and it serves (the data physically belongs to the person, the architecture does not permit any other arrangement). A continuity layer that satisfies the first constraint without the second is not a successful product. It is a hazard.

\section{What this is not}

It is worth being precise about what this thesis does not claim, because the precision is what makes the thesis defensible.

It does not claim that models will stop mattering. They keep mattering. They are the processor. Processors keep mattering even after the storage layer becomes the durable thing.

It does not claim that the continuity layer is a competitor to OpenAI or Anthropic. It is not. Those companies build the brain. The continuity layer makes the brain remember it is alive. They are different layers and they will eventually need each other.

It does not claim that the layer is finished. The first reference implementation passes the benchmark today. The road from ``passes a benchmark'' to ``becomes the standard underneath every AI system'' is long, and it requires research, capital, distribution, and time. The thesis is not ``we are done.'' The thesis is that the inevitability of this layer is now visible, and that the work of building it has begun in public.

It does not claim that continuity is hard because of compute. It is hard because nobody has built the right primitive. The compute requirement is small. The conceptual requirement, building a storage system whose unit is a reconstructed situation and not a row, is what makes the work hard, and it is what makes most existing attempts fall short of the seven properties.

It does not claim that the four-layer arc is finished. Layer 1 exists. Layer 2 is the next research round. Layer 3 follows from layer 2 on a timescale measured in years. Layer 4 follows from layer 3 on a timescale measured in decades. Each step is a derivation from the previous one, but each step is real work, and none of it is done in advance.

It does not claim that this is the only thing that matters in AI infrastructure. There are other layers that need to exist. Continuity is the first one that has both a clear definition and a working reference implementation. That is enough to argue for it as a category. It is not enough to argue that nothing else matters.

It does not promise that the technology will solve AGI, end forgetting forever, or change the world by Tuesday. The work is what the work is. The reader can draw the implications.

\section*{Closing}

The question this essay opened with was: what does intelligence accumulate against?

For most of computing history, the answer was: storage. State accumulated against files, against rows, against documents, against blobs. The intelligence happened somewhere else and the storage stayed still.

AI inverted that arrangement without naming the inversion. The intelligence is now the part that stays still, frozen weights released and replaced on a cadence. The state is the part that disappears, every time a session ends, with nothing carried forward.

The continuity layer is what restores the older arrangement, in a form that fits the new stack. The state becomes persistent again. The intelligence becomes the part that runs against it. The layer that holds the state, the thing that compounds, that resists commoditization, that becomes more valuable over time without additional capital, becomes the durable thing in the system.

Whether or not anyone funds this work, the layer is going to exist. The physics wall will force it. The economics will force it. The experience of using AI products that forget you between sessions will force it. The only open questions are who builds it, how it gets defined, and whether it gets defined in a way that preserves the people it carries forward.

The work is in public. The benchmark is published. The reference implementation passes it. The architecture is named, and the name is the architecture. The mark is the property the architecture produces. The layer is the company. The model is the processor. The layer is what stays.

\bibliographystyle{plainnat}
\bibliography{references}

\end{document}